\documentclass[sigconf,nonacm]{acmart}

\AtBeginDocument{%
  \providecommand\BibTeX{{%
    \normalfont B\kern-0.5em{\scshape i\kern-0.25em b}\kern-0.8em\TeX}}}

\setcopyright{none}
\copyrightyear{}
\acmYear{}
\acmDOI{}

\acmBooktitle{} 
\acmPrice{}
\acmISBN{}

\begin{document}

\title{Outfit Completion via Conditional Set Transformation}


\author{Takuma Nakamura}

\affiliation{%
  \institution{ZOZO Research}
  \city{Tokyo}
  \country{Japan}}

\author{Yuki Saito}
\affiliation{%
  \institution{ZOZO Research}
  \city{Tokyo}
  \country{Japan}}
\email{yuki.saito@zozo.com}

\author{Ryosuke Goto}
\affiliation{%
  \institution{ZOZO Research}
  \city{Tokyo}
  \country{Japan}}
\email{ryosuke.goto@zozo.com}


\begin{abstract}
In this paper, we formulate the outfit completion problem as a set retrieval task and propose a novel framework for solving this problem. The proposal includes a conditional set transformation architecture with deep neural networks and a compatibility-based regularization method. The proposed method utilizes a map with permutation-invariant for the input set and permutation-equivariant for the condition set. This allows retrieving a set that is compatible with the input set while reflecting the properties of the condition set. In addition, since this structure outputs the element of the output set in a single inference, it can achieve a scalable inference speed with respect to the cardinality of the output set. Experimental results on real data reveal that the proposed method outperforms existing approaches in terms of accuracy of the outfit completion task, condition satisfaction, and compatibility of completion results.

\end{abstract}




\maketitle

\section{Introduction}
Automating outfit generation is a promising technology with a variety of fashion-related applications. In addition to saving time in daily outfit selection, it has the potential to improve personalized recommendation systems on fashion e-commerce and support stylists' outfit suggestions. The term "outfit" here refers to a combination of garments that are intended to be worn together. This concept is not merely a combination of complementary garments. Actual outfits include information such as the place and situation where they are worn, the season and climate, as well as fashion style and inner expression, and there is a complex interaction that goes beyond mere color and pattern compatibility.

In the research area of computer vision, there has been active discussion on data structures and algorithms for learning the goodness of such outfits. Various methods have been proposed and successfully applied to quantify the compatibility of outfits. \cite{cucurull2019context, saito2020exchangeable, han2017learning}. However, most of these methods focus on evaluating complete outfits and cannot directly solve the problem of completing partially missing outfits by selecting a set of garments. In these frameworks, the computational cost of outfit completion is high because it requires iterative updates to optimize combinations that increase the compatibility score.

\begin{figure}[t]
  \centering
  \includegraphics[width=\linewidth]{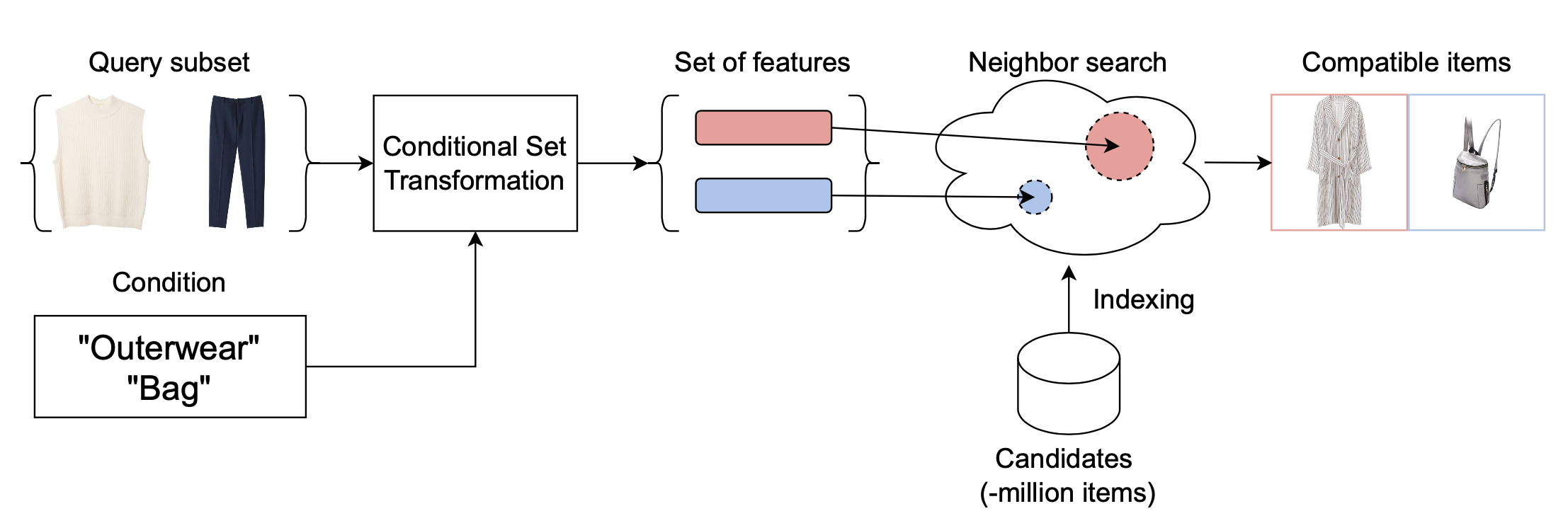}
  \caption{A schematic view of the architecture for solving the outfit completion problem as a conditional set retrieval task. The red and blue squares represent features transformed from the "Outerwear" and "Bag" labels, respectively. By performing neighborhood search for each feature, the method can identify missing items in the outfit that are compatible with the input set.}
  \label{fig:schematic}
\end{figure}

In this paper, we formulate the outfit completion problem as a conditional set retrieval task (Figure \ref{fig:schematic}). set retrieval does not require iterative optimization and directly predicts a set that is compatible with the query set. Moreover, set retrieval enables real-time inference using approximate nearest neighbor algorithms (e.g. ScaNN \cite{guo2020accelerating}, NGT \cite{ iwasaki2018optimization}) even when the search space is large.

Traditional set retrieval tasks often assume homogeneous instances, where the goal is to find a similar set of instances from the gallery \cite{zaheer2017deep}. However, the outfit completion task in fashion poses unique challenges because it requires dealing with heterogeneous instances. Unlike a standard set retrieval, where all instances belong to the same category, fashion outfits consist of various types of garments as tops, bottoms, and accessories. This heterogeneity makes it non-trivial to calculate compatibility scores or find the most compatible set. Our approach is specifically tailored to handle these complexities.

To solve the set retrieval problem, we propose a conditional set transformation architecture using deep neural networks (DNNs) and a compatibility-based regularization training method. The proposed transformation incorporates an attention mechanism \cite{vaswani2017attention} that takes into account the relationships between set elements. Additionally, an extension to a conditional model enables control over the attributes of the output set. By training this model with compatibility-based regularization, we achieve retrieval of a set that is compatible with the query set.

We evaluate the proposed method through experiments using a large dataset collected via a real-world web service. In the outfit completion experiments, our method successfully and recovers the original outfit with high accuracy and satisfies the input conditions compared to existing methods, and scales efficiently in terms of inference time as the number of missing elements increases. Human evaluation indicate that in 34\% of test cases, our method generates better outfits than those created by humans.

\section{Preliminaries}
\subsection{Conditional Set Transformation}

We introduce the following necessary notation. Let $\bm{x}_n,\bm{y}_m,\bm{z}_m \in \mathfrak{X} = \mathbb{R}^D$ be feature vectors, and the sets of these vectors are denoted as $X=\{\bm{x}_1,...,\bm{x}_{N}\}$, $Y=\{\bm{y}_1,...,\bm{y}_{M}\}$, and $Z=\{\bm{z}_1,...,\bm{z}_{M}\}$, where $X, Y, Z\in 2^\mathfrak{X}$.

We introduce the transformation $f:2^\mathfrak{X} \times 2^\mathfrak{X} \rightarrow 2^\mathfrak{X}$. We assume that the transformation is permutation-invariant with respect to X and permutation-equivariant with respect to Y \cite{zaheer2017deep}. Here, permutation-invariant implies that the permutation of $X$ does not affect the permutation of $Y$, while permutation-equivariant implies that the permutation of $Y$ changes according to the permutation of $Z$. Specifically, for any permutation $\pi$, $Y = f(\pi X, Z)$ and $\pi Y = f(X, \pi Z).$ In this paper, we refer to the above transformation as a conditional set transformation and use it for the set retrieval task.

\subsection{Outfit Completion}
An outfit is a set of complementary items that can be worn at the same time, such as a combination of a shirt and a skirt. Let $X$ and $Y$ be subsets of an outfit and $x_n$ and $y_m$ be elements of each subset, with $X \cap Y = \emptyset$.

Let $z_m$ be an attribute of $y_m$\textemdash for example, category information such as tops or bottoms. We define the outfit completion task as predicting $Y$ using $X$, $Z$, and conditional set transformation $f$

\begin{equation}
    Y = f(X,Z).
\end{equation}

In this task, $f$ is optimized by training the following constraints: $X$ and $Y$ are complementary to each other, and each element of $Y$ reflects the attributes specified by $Z$.

\subsection{Set Compatibility}
Compatibility is a concept introduced to evaluate how the combination of $X$ and $Y$ constitutes an outfit, which can be quantified by the Set Matching model \cite{saito2020exchangeable}. Let $s \in \mathbb{R}$ be a compatibility score and let $g$ be a Set Matching model; then,

\begin{equation}
s = g(X,Y).
\end{equation}

Here, $g : 2^\mathfrak{X} \times 2^\mathfrak{X} \rightarrow \mathbb{R}$, and $X$ and $Y$ are exchangeable.

\subsection{Slot Attention}
Slot Attention \cite{locatello2020object} was proposed as a permutation-equivariant architecture for obtaining a set with an arbitrary number of elements. The $l$-layer transformation is

\begin{equation}
    Y_{l+1} = SlotAttention_l(Y_l, X),
\end{equation}

where the first layer $Y_0$ is initialized with a Gaussian distribution.

Slot Attention contains  Multi-Head Attention Blocks (MAB)\cite{vaswani2017attention} and is permutation-invariant with respect to $X$ and permutation-equivariant with respect to $Y_0$.

Since the output of Slot Attention is a set, a permutation-invariant loss function must be used for training. The Chamfer loss function \cite{kosiorek2020conditional} is one such function.

Given the distance between $x$ and $y$ as $d(x,y)$, we have

\begin{equation}
\label{chamferloss}
\mathcal{L}_\mathrm{Chamfer}(X,Y) = \sum_{x \in X} \min_{y \in Y} d(x,y) + \sum_{y \in Y} \min_{x \in X} d(x,y).
\end{equation}

\section{Methodology}
\begin{figure*}
    \centering
    \includegraphics[width=0.9\textwidth]{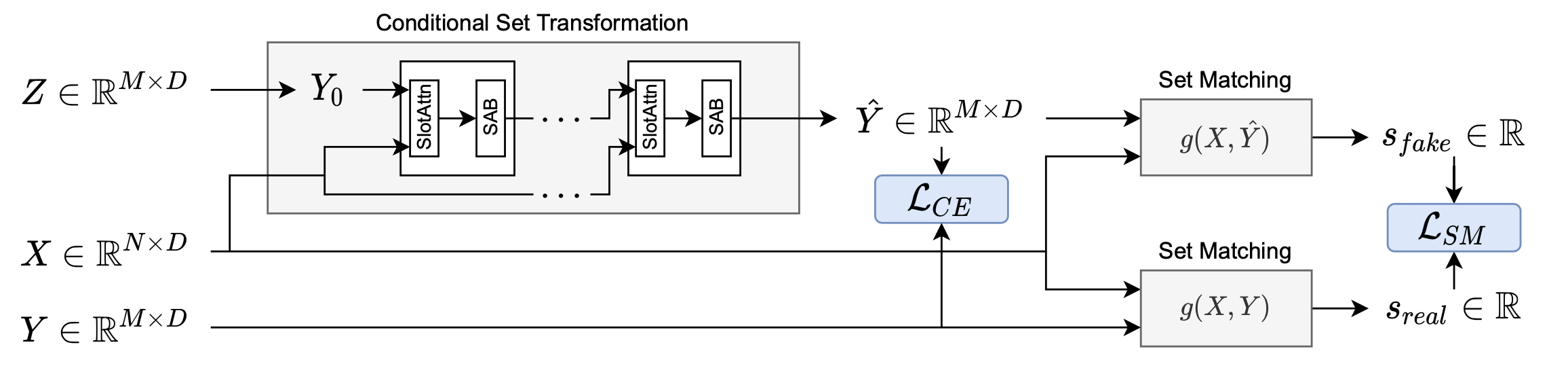}
    \caption{Overview of the proposed network. The proposed network architecture comprises of two key components: (1) A Conditional Set Transformation (CST) module (described in Section 3.1) that generates a set of features $\hat{Y}$ from a given query set $X$ and a conditional set $Z$, and (2) a Set Matching module (described in Section 3.3), which quantifies the compatibility between the two sets.}
    \label{fig:fig2}
\end{figure*}

The model for the outfit completion problem is presented in Figure \ref{fig:fig2}.


\subsection{Conditional Set Transformation Model}
To control the attributes of each element in the output set $\hat{Y}$ with the condition set $Z$, we propose replacing the initial set $Y_0$ of Slot Attention with $Z$. For example, in outfit completion, consider the case in which we specify a category for each item in $\hat{Y}$. Since Slot Attention is permutation-equivariant with respect to $Y_0$, the order of elements in the condition set $Z$ and the output set $\hat{Y}$ correspond to each other.

In addition, if the goal is also to make the output set itself compatible, the relationships among the elements cannot be ignored. To address this problem, we introduce a Set Attention Block (SAB) \cite{lee2019set}, a set transformation based on the relationships between elements, to transform the result of Slot Attention. The final transformation is

\begin{equation}
\hat{Y} = SAB(SlotAttention(Z,X)).
\end{equation}

Since the SAB is permutation-equivariant to its inputs, the above transformation is permutation-invariant for $X$ and permutation-equivariant for $Z$, as described in Section 2.1. By stacking these modules, we can achieve complex transformations while preserving the order of conditions.

\subsection{Prediction Evaluation}

The set transformation was evaluated using a permutation-invariant loss function \cite{kosiorek2020conditional}, as the order of the output set $\hat{Y}$ cannot be fixed. On the other hand, in the proposed conditional set transformation, the output set is permutation-equivariant with respect to the condition set and, thus, the order of $\hat{Y}$ can be controlled by the condition set. In other words, the order of $\hat{Y}$ and $Y$ can be easily mapped. Therefore, there is no need to consider the permutation-invariant of the loss function, and the elements of the output set can be evaluated individually using the cross-entropy loss. The loss between model prediction $\hat{Y}$ and the ground truth $Y$ is given by

\begin{equation}
\label{crossentropyloss}
    \mathcal{L}_{CE}(\hat{Y},Y) = -\frac{1}{M}\sum^M_{m=1}\log \frac{\exp (\hat{y}^{T}_{m}\cdot y_m)}{\sum_{y \in \mathcal{Y}}\exp (\hat{y}^{T}_{m}\cdot y_m)},
\end{equation}

where $\mathcal{Y}$ is the set of all items. Usually, the computation for $\mathcal{Y}$ is expensive; thus it is replaced by all items in the mini-batch at training phase \cite{sohn2016improved}.

\subsection{Set Matching Regularization}
In the outfit completion problem, the goal is to make output $\hat{Y}$ compatible with a pairwise set $X$. When $(X,Y)$ is a pair of ground truth, $\hat{Y}$ is evaluated with the following softplus function:

\begin{equation}
\label{setmatchingloss}
    \mathcal{L}_{SM}(\hat{Y},X,Y) = \log (1 + \exp (g(X,Y) - g(X,\hat{Y}))),
\end{equation}

where $(X,Y)$ is assumed to be a manually created outfit. 
The less compatible the pair $(X, \hat{Y})$ is compared to the ground truth pair $(X,Y)$, the greater the penalty to the model. The effectiveness of the regularization in the outfit completion problem depends on the reliability of the Set Matching model, which is confirmed in Appendix \ref{appendix:validation}.

\subsection{Training Strategies}
The loss function in the outfit completion problem is as given below:

\begin{equation}
\label{totalloss}
    \mathcal{L}(X,Y,Z) = \mathcal{L}_{CE}(CST(X,Z;\theta),Y) + \alpha \mathcal{L}_{SM}(CST(X,Z;\theta),X,Y),
\end{equation}

where $CST()$, $\theta$, and $\alpha$ are the conditional set transformation model, its parameters, and regularization weight, respectively. Since the Set Matching model is used as a loss function, its parameters are pre-trained and fixed when training the conditional set transformation. In addition, the parameters of the feature extractor are fixed as well. Alternatively, the Set Matching model can be trained simultaneously with the feature extractor, as proposed in \cite{saito2020exchangeable}.  All training is performed using stochastic gradient descent (SGD).

\section{Related work}
\textbf{Fashion Outfit Compatibility}
Fashion outfit recommendation systems have evolved significantly over the years. Early systems like Iwata el al. \cite{iwata2011fashion} and Liu et al. \cite{liu2012hi} focused on recommending fashion items in pairs. They used methods such as probabilistic topic models and attribute extraction to learn visual relationships between different fashion items. However, these systems only recommend for predefined category pairs and cannot directly handle outfits with more than three items. To deal with real-world outfits, it is desirable to be able to handle variable-length inputs. Han et al. \cite{han2017learning} and Nakamura and Goto \cite{nakamura2018outfit} used bidirectional LSTM to learn sequences of fashion items, treating outfits as ordered categories. Saito et al. \cite{saito2020exchangeable} explored set-to-set matching in fashion outfits, defining a task to measure the compatibility of different item sets using exchangeable deep neural networks. Cucurull et al. \cite{cucurull2019context} used graph structures to represent relationships between items and applied graph neural networks (GNNs) to predict fashion compatibility.

While these methods made great strides by allowing direct input of variable length inputs, they also had limitations. They required iterative computations to optimize the scoring function to fill in the missing garments. A discussion of the computational complexity of the proposed method and related outfit completion methods is included in the Appendix \ref{appendix:complexity}. Our research aims to directly address these gaps by developing a system that can efficiently and accurately complete partial missing outfits.We contribute to the field by providing a more holistic and practical solution to the fashion outfit task.

\textbf{Neural Set Functions} Neural Set Functions are machine learning methods that treat sets as input. The most notable models include the Deep Sets \cite{zaheer2017deep} and the Set Transformer \cite{lee2019set}. Deep Sets aggrigate features across instances to learn an order-independent representation neural networks. Set Transformer, on the other hand, uses a self-attention mechanism to capture higher-order relationships between instances.

In addition, the Conditional Neural Processes \cite{kosiorek2020conditional} and Object-Centric Learning \cite{locatello2020object} have been proposed as set transformation models. These methods are used for point cloud reconstruction and arbitrary number of objects detection. However, these methods are designed for homogeneous elements. An outfit is a combination of different categorical elements and requires a different approach than point cloud reconstruction or arbitrary number of objects detection. In this study, we treat outfit as a set and aim to apply these technologies to real-world situations. Specifically, we develop a set transformation model to recognize heterogeneous instances (e.g., different types of garments) and understand their relationships. This approach allows us to develop models that capture the complexity and diversity of fashion data and efficiently complete partially missing outfits.

\section{Experiments}

\subsection{Problem Setting}
The query and target sets are obtained by randomly splitting a human-created outfit into two parts. We designed a task to predict the items of the target set from the query set and the category information of the target set. The proposed method predicts the items of the output set in a single inference and evaluates each of them separately. For comparison, we conducted experiments with the six methods listed in Table \ref{tab:methods}, including the proposed model. CR and Cx, which input a condition set, use a lookup table to convert category labels into a condition set. For xR, xx, and sa \textemdash which do not use a condition set \textemdash a mask to specify the number of elements is used instead of category information, and a normal random number is used as the initial value of Slot Attention. 

\begin{table*}[t]
  \caption{List of methods for the ablation study. The proposed method is denoted as CR because it introduces the extension to the Conditional model and the Regularization. Ablation studies are denoted by the symbol x, which represents the missing part of the model. Abbreviations for ablation studies are denoted by Cx, xR, and xx. sa and st are acronyms of the existing methods.}
  \label{tab:methods}
  \begin{tabular}{ccc}
    \toprule
    Symbols &Methods &Loss Functions \\
    \midrule
    CR & Conditional set transformation (proposed method) & eqn.\ (\ref{totalloss})\\
    Cx & Set Matching regularization removed from CR& eqn.\ (\ref{crossentropyloss}) \\
    xR & Set Matching regularization added to xx& eqn.\ (\ref{chamferloss}) + eqn.\ (\ref{setmatchingloss})\\
    xx & SAB added to sa & eqn.\ (\ref{chamferloss})\\
    sa & Slot Attention \cite{locatello2020object} & eqn.\ (\ref{chamferloss})\\
    st & Set Transformer \cite{lee2019set} & eqn.\ (\ref{crossentropyloss})\\ 
  \bottomrule
\end{tabular}
\end{table*}

The SHIFT15M dataset \cite{kimura2021shift15m} is used for training and evaluation. This dataset includes approximately 2.56 million outfits posted on the fashion outfit application IQON. Each outfit in the dataset contains information like a list of items and the number of LIKEs (votes from users on the IQON). Additionally, each item is associated with its category and image features. For this experiment, we constructed a dataset for training and evaluation by extracting 583,788 outfits that received a number of LIKEs in at least the 75th percentile and at least five items per outfit.

\subsection{Evaluation}

The conditional set transformation produces a set of features of items that complement the query set. As a retrieval task, we evaluate whether or not each element of the output set and the target set match at the item level. The quality of the ouput set itself is also evaluated using a trained Set Matching model.

\textbf{Recall}: An item database was constructed by extracting the features of approximately 260,000 items from the test data. A nearest neighbor search was performed using the features of the output set as queries, and 32 similar items were selected from the database. When the number of elements in the output set was $M$, we obtained $32M$ candidate items per outfit. Recall is calculated between this and the list of items in the target set.

\textbf{Accuracy}: The category list of the output set is compared with the category list of the target set and evaluated in terms of accuracy. The higher the accuracy, the more correctly the condition is reflected.

\textbf{Set Matching Score Difference (SMD)}: The pairs of the query set and the output set were input to the trained Set Matching model and corresponding scores were obtained. The same evaluation was performed for the query set and target set pairs and the difference of the two is reported as the Set Matching Score Difference (SMD).

Recall is used to evaluate whether an appropriate set of items can be retrieved for a query set, and Accuracy is used to evaluate whether the retrieval results satisfy the necessary conditions. On the other hand, these do not take into account the matching quality between the query set and the output set, and SMD is used to compensate for this aspect. Thus, it can be said that the higher all the indices are, the better the outfit completion method.

\subsection{Quantitative Results}
\begin{figure}[h]
  \centering
  \includegraphics[width=\linewidth]{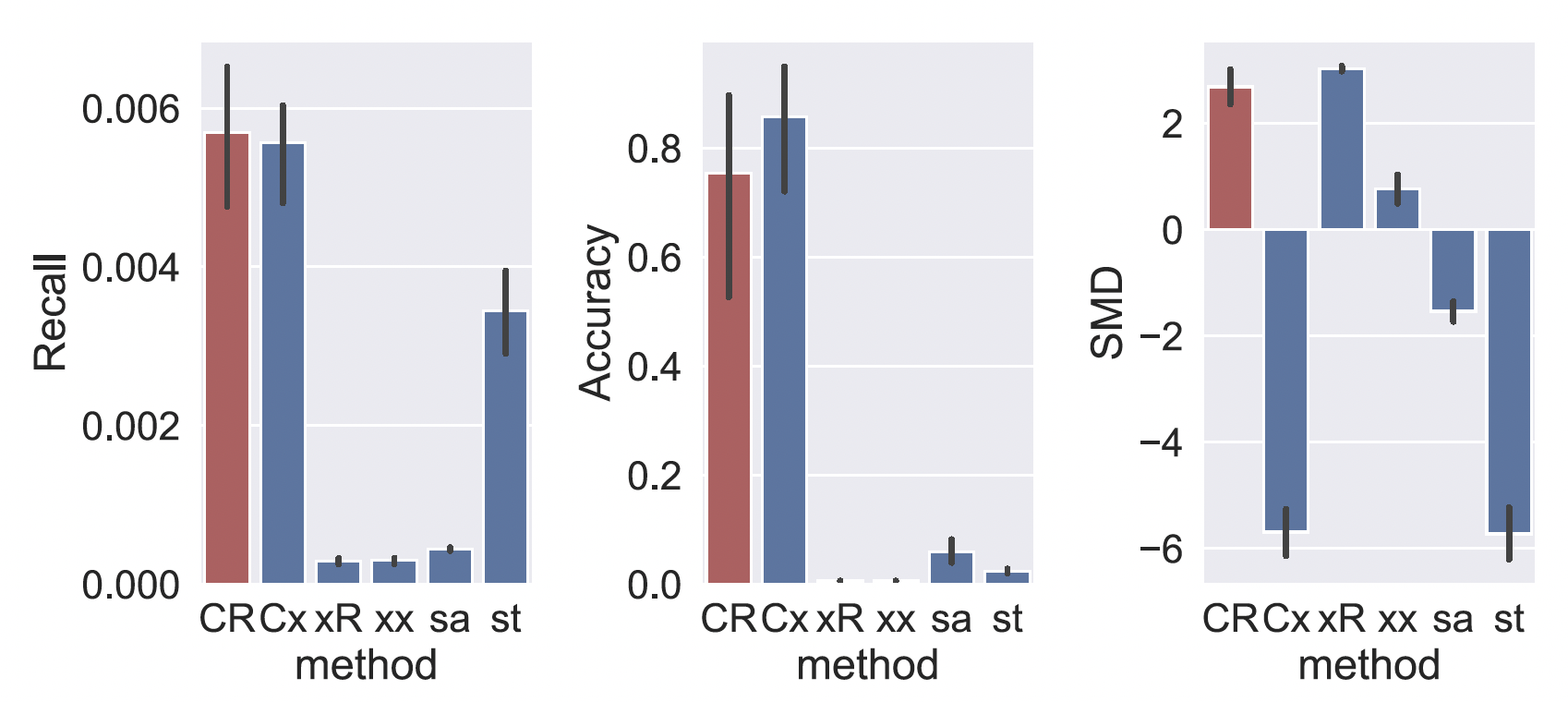}
  \caption{Quantitative evaluation results.}
  \label{fig:evaluation}
\end{figure}

The results of the quantitative evaluation are depicted in Figure \ref{fig:evaluation}. CR and Cx outperform the other methods in Recall and Accuracy, thereby confirming the contribution of the conditional model to the set retrieval task. The adoption of the cross-entropy loss function also contributes to the improved performance. Since item selection is essentially a discriminative problem, cross-entropy loss is a more natural formulation than Chamfer loss and is considered more appropriate for the outfit completion problem. Further, the values of SMD were higher for CR and xR than for the other methods, thereby confirming that the Set Matching regularization contributes to compatibility. In particular, the SMDs of these methods are above 0, thereby indicating that the outfits completed by these methods are, on average, more compatible than those created by humans in the Set Matching criterion.

\subsection{Qualitative Results}
\begin{figure*}
  \centering
  \includegraphics[width=\linewidth]{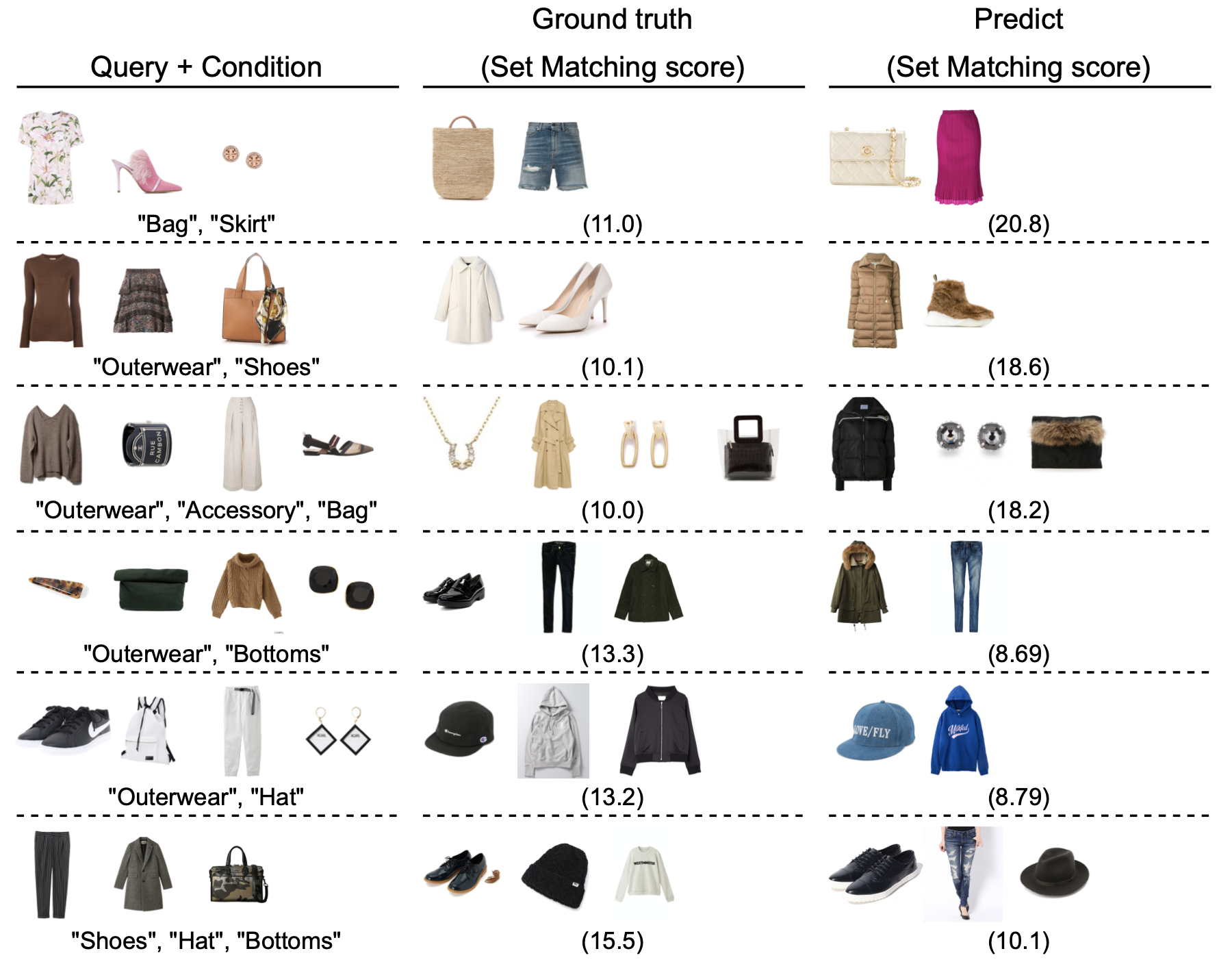}
  \caption{Examples of completion results. The left column indicates the item images in the query set and the category labels in the condition set. The middle column depicts the item image of the target set (ground truth). Below the images are the Set Matching scores combined with the query set. The right column depicts the nearest neighbor items determined by the output of the proposed method, and their corresponding Set Matching scores are also displayed at the bottom.}
  \label{fig:example}
\end{figure*}

Figure \ref{fig:example} depicts examples of outfit completion results. We define a good retrieval result as one in which the query set and the prediction are naturally compatible, and the category of the item in the prediction matches the category label of the condition. The proposed method satisfies the input conditions and is able to select a set of items that are consistent in terms of outfit. In particular, it was confirmed that the proposed method makes compatible predictions in terms of color and season.

Further, consistency was also observed across fashion styles, such as casual and sports. We note that such abstract concepts become more apparent when multiple items are combined. Interpreting the results from this perspective, it is possible that the proposed method performs inference by considering not only the image features of the items but also the features of the combinations.

The completion example in the fifth row illustrates that while the ground truth contains two outerwear items, the prediction suggests only one. This is because the proposed method can only generate one feature for each category and fails to select items that are different from each other when the categories overlap. To achieve a more generalized completion, one possible direction for future research is to develop a set transformation that allows for overlapping categories and items.

\subsection{Human Evaluation}
The effectiveness of the Set Matching Score Difference (SMD) metric depends on the performance of the Set Matching Model. To eliminate this dependency, we conducted a human evaluation experiment as an independent measure. Seven female fashion professionals were recruited as evaluators and they were presented with both the ground truth and the results generated by the proposed method, with the answers hidden. They were asked to select the superior result in terms of outfit compatibility. The results are detailed in Table \ref{tab:humanevaluation}. The number of valid votes was 984, and the rate of selecting the proposed method did not differ significantly among the evaluators. On average, the proposed method was considered superior to the ground truth in $34\%$ of the test cases. While the results generated by the proposed method do not consistently outperform the ground truth, it is confirmed that the proposed method yields superior generation performance in several cases.

\begin{table}[t]
  \caption{Human evaluation results.}
  \label{tab:humanevaluation}
  \begin{tabular}{cccc}
    \toprule
    Evaluator ID & \# of votes & \begin{tabular}{c}\# of proposed \\ method voted \end{tabular} & rate \\
    \midrule
    1 & 282 & 93 & $0.329 (\pm 0.054)$ \\
    2 & 189 & 63 & $0.333 (\pm 0.067)$\\
    3 & 124 & 45 & $0.362 (\pm 0.084)$\\
    4 & 157 & 53 & $0.337 (\pm 0.073)$\\
    5 & 83 & 27 & $0.325 (\pm 0.100)$\\
    6 & 58 & 23 & $0.396 (\pm 0.125)$\\ 
    7 & 91 & 34 & $0.373 (\pm 0.099)$\\ 
  \bottomrule
    total & 984 & 338 & $0.343 (\pm 0.029)$\\ 
  \bottomrule
\end{tabular}
\end{table}

\subsection{Inference Time Comparison for Outfit Completion}

\begin{figure}[h]
  \centering
  \includegraphics[width=\linewidth]{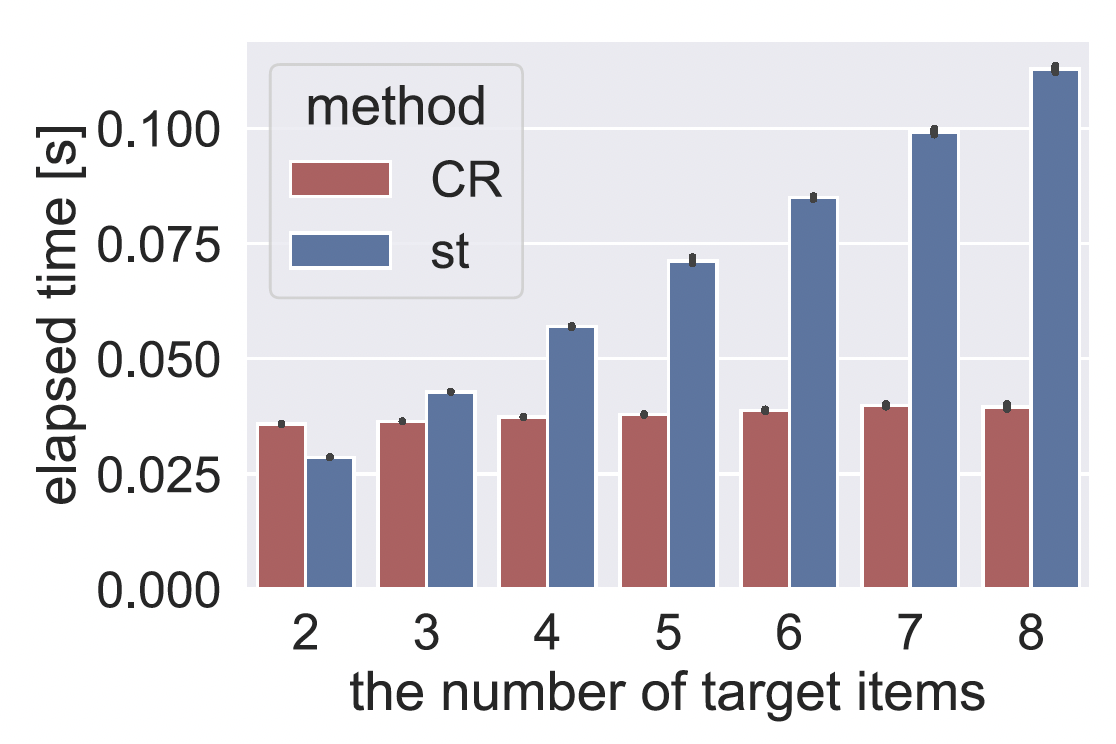}
  \caption{Inference time of the proposed and compared method.}
  \label{fig:elapsed_time}
\end{figure}

The comparison between the proposed method and the set transformer in terms of outfit completion time is presented in Figure \ref{fig:elapsed_time}. The elapsed time for the Set Transformer increases with the number of elements to be completed, whereas the proposed method completes all elements in a single inference, resulting in a time that remains almost constant. The proposed method is more efficient than the Set Transformer and more scalable to the number of elements to be completed. This result is consistent with the computational complexity analysis in Appendix \ref{appendix:complexity}. Consequently, the proposed method has the potential to meet the requirements of applications that require immediate responses, such as product recommendation in e-commerce.

\section{Conclusion}
In this paper, we formulate a set retrieval task for the outfit completion problem and propose a conditional set transformation and its training method. Experiments on real data reveal that the proposed method can predict a set that is compatible with the input set and can control the attributes of the elements of the output set. We also find that the completed outfits have compatibility in terms of colors, seasons, and fashion styles. In human evaluation experiments, the completion results of the proposed method were judged to be better than the ground truth in 34\% of the test cases. The proposed method was also found to be scalable in terms of inference time as the number of missing elements increases.

\bibliographystyle{ACM-Reference-Format}
\bibliography{sample-authordraft}

\appendix

\section{Computational Complexity of Outfit Completion}
\label{appendix:complexity}

The completion of missing items in the outfit completion problem entails two steps:
\begin{itemize}
    \item[] \textbf{Step 1} \textemdash Generating features for the items to be completed using a DNN, with computational complexity denoted as $P$.
    \item[] \textbf{Step 2} \textemdash Selecting the items via nearest neighbor search based on the generated features, with computational complexity denoted as $Q(N_y)$, where $N_y$ is the number of candidate items.
\end{itemize}

It should be noted that $N_y$ is typically very large in practical applications and, consequently, approximate nearest neighbor algorithms, such as ScaNN \cite{guo2020accelerating}, are often utilized to facilitate candidate selection. Therefore, the actual time required for $Q(N_y)$ is assumed to be limited to a few tens of milliseconds.

The following are the computational complexities of the proposed method, Set Transformer \cite{lee2019set}, and BiLSTM \cite{han2017learning} when applied to the outfit completion problem:

\textbf{Proposed Method}: The proposed method simultaneously generates features for multiple items in a single inference, and then performs nearest neighbor search for each feature individually. When the number of items to be completed is $M$, the computational complexity of the proposed method is $P + M \cdot Q(N_y)$.

\textbf{Set Transformer}: When applied to the outfit completion problem, this algorithm sequentially adds missing items. The computational complexity is $M \cdot (P + Q(N_y))$.

\textbf{BiLSTM}: The series generation algorithm detailed in Sect. 5.5 of BiLSTM \cite{han2017learning} performs the same number of inferences and nearest neighbor searches as the series length as well as execute the same number of iterations as the size of the query set. Therefore, when the series length is $L$, the computational complexity is $(L-M)\cdot L\cdot (P+Q(N_y))$.

Set Transformer and BiLSTM realize set completion through repeated element selection. To achieve high-quality completion under these constraints, it is desirable to use them in combination with optimization algorithms, such as beam search. However, the large inference time required for iterations may impede real-world applications.

\section{Additional Experiments}
\subsection{Validation of Set Matching Model}
\label{appendix:validation}
\begin{figure}[h]
  \centering
  \includegraphics[width=\linewidth]{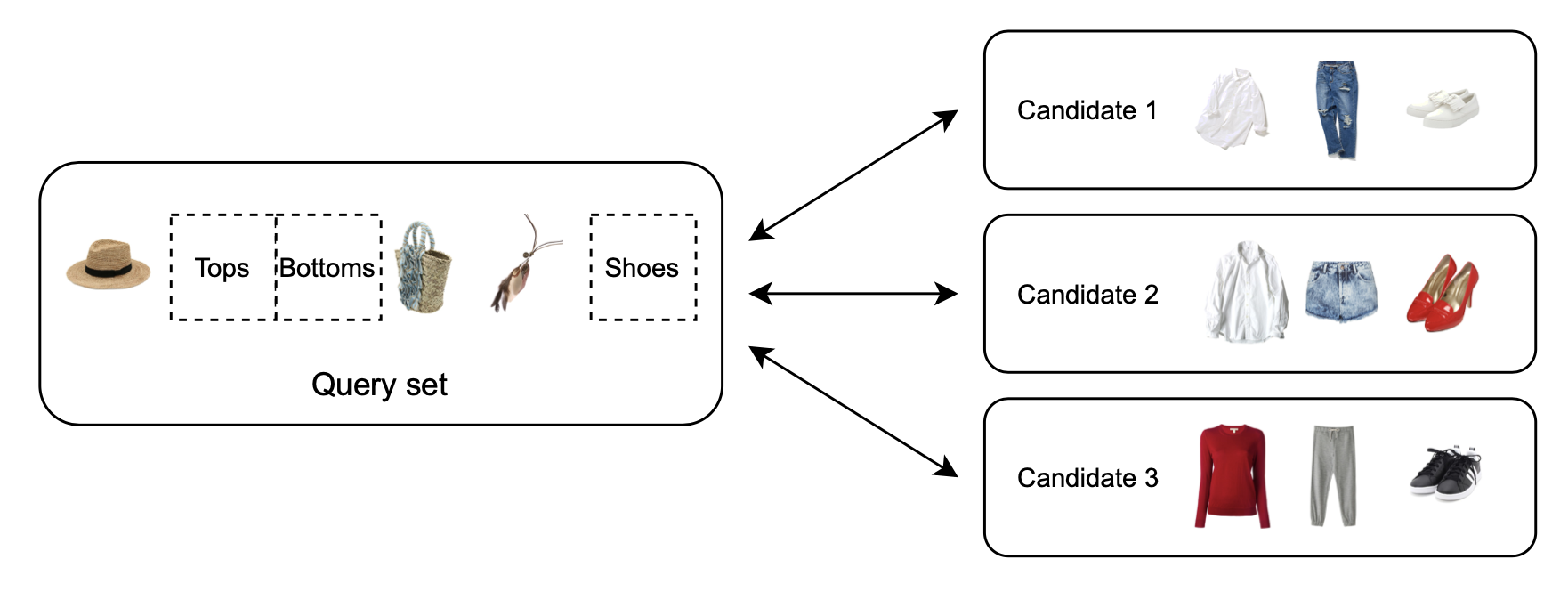}
  \caption{Examples of FINB task}
  \label{fig:finb}
\end{figure}

We evaluate the accuracy of the Set Matching model, which regularizes and calculates compatibility scores in the proposed method, using the Fill-In-The-N-Blank (FINB) task as well as \cite{saito2020exchangeable}. The FINB task uses a subset of outfits as the query set and selects missing items from the candidates (Figure \ref{fig:finb}). Unlike set retrieval, the FINB task requires the preparation of positive and negative examples in advance. The query set and positive example pairs were created by randomly splitting manually generated outfits, and the negative examples were randomly selected from a set of items in the same category as the positive examples. For each of the 58,379 outfits included in the test data, seven negative examples were created and eight choice discrimination questions were constructed.

\begin{table}
  \caption{Results of the FINB task.}
  \label{tab:accuracy}
  \begin{tabular}{ccc}
    \toprule
    Methods & Accuracy \\
    \midrule
    CR (proposed method) & $0.426(\pm 0.027)$\\
    Set Matching & $0.624(\pm 0.003)$ \\
  \bottomrule
\end{tabular}
\end{table}

The proposed method and the Set Matching model were evaluated in terms of their accuracy or the ability to select the correct candidate. Both methods have different candidate selection procedures, as explained in the following paragraph.

The proposed method calculates the similarity between the features of missing items and the items included in the candidates individually, and then selects the candidate with the largest sum of similarities. The Set Matching model calculates Set Matching scores for each candidate and selects the candidate with the largest score. The results are presented in Table \ref{tab:accuracy}. These results demonstrates that the Set Matching model has reliable performance and can provide better outfit completion than the proposed method when the search space is limited.

\subsection{Distribution of Set Matching Scores}
\begin{figure}[h]
  \centering
  \includegraphics[width=\linewidth]{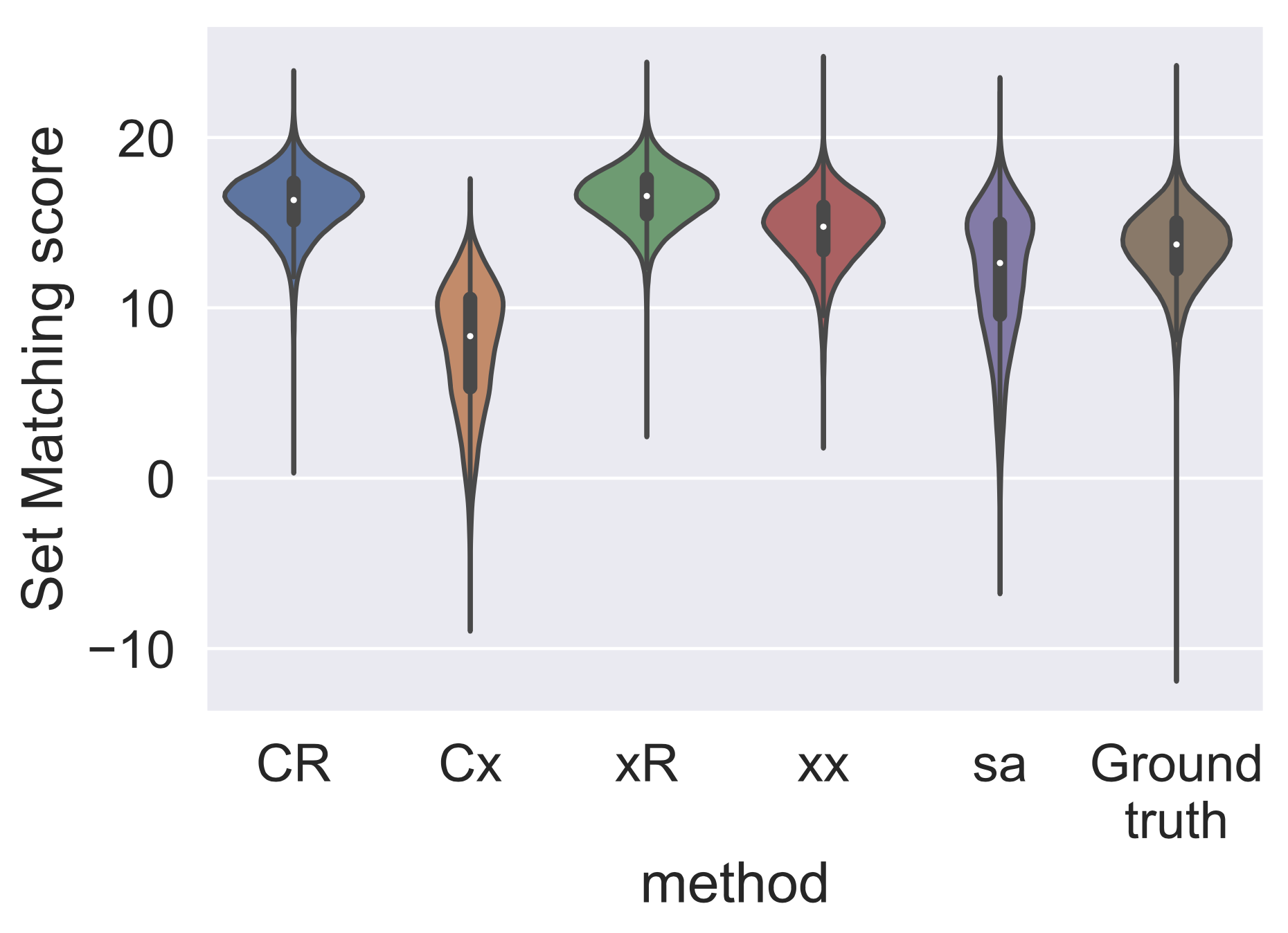}
  \caption{Distributions of Set Matching Scores.}
  \label{fig:fig6}
\end{figure}

Figure \ref{fig:fig6} illustrates the distribution of Set Matching scores for outfits completed by each method on the test data. We observed that the Set Matching scores for outfits stored with Cx without Set Matching regularization tended to be inferior to those of the ground truth. The extension of the conditional model and the implementation of cross-entropy significantly enhanced retrieval performance, although their impact on the Set Matching score cannot be definitively confirmed. This suggests that Set Matching regularization can be used to achieve natural completion.

\subsection{Diversity of Outfit Completion Results}
\begin{figure}[h]
  \centering
  \includegraphics[width=\linewidth]{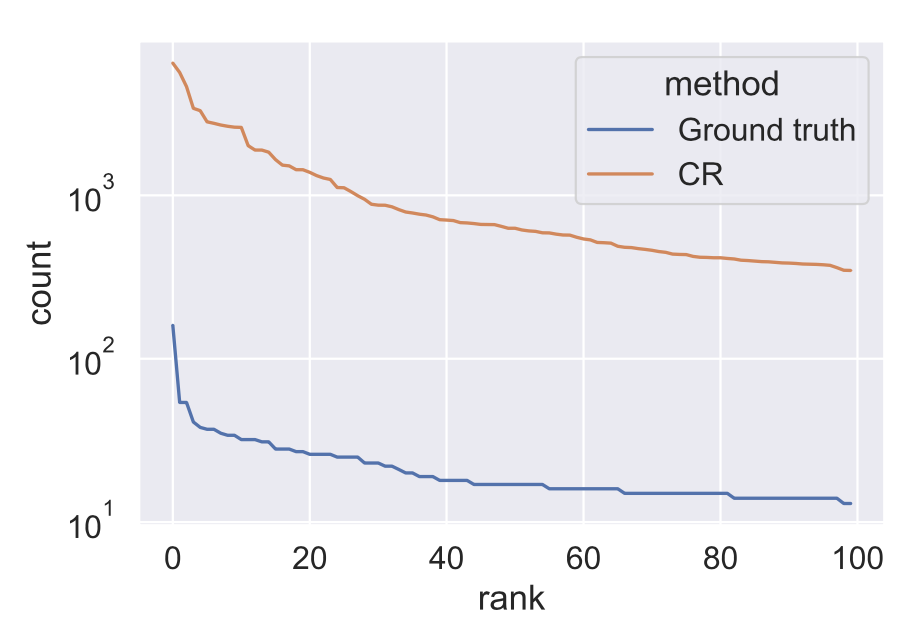}
  \caption{Frequency distribution of items selected for completion.}
  \label{fig:frequency}
\end{figure}

Using the proposed method, we analysed how often each item is selected when completing outfits. The test data consists of 258,417 candidate items. The number of times that each item was used in approximately 58,000 outfit completions was recorded. The top 100 items in order of frequency were plotted as a histogram (Figure \ref{fig:frequency}). A clear discrepancy was observed between the proposed method's item selection tendencies and the ground truth. The highest ranked item in the ground truth appeared only approximately 160 times, and item usage was relatively evenly distributed. In contrast, the completions of the proposed method were heavily skewed toward certain items. Thus, it can be concluded that the proposed method does not provide diverse completions. Addressing this issue is a task for future research.

\end{document}